\def\year{2021}\relax
\documentclass[letterpaper]{article} 
\usepackage{aaai21}  
\usepackage{times}  
\usepackage{helvet} 
\usepackage{courier}  
\usepackage[hyphens]{url}  
\usepackage{graphicx} 
\usepackage{graphicx}  
\usepackage{natbib}  
\usepackage{caption} 
\title{Learning to Cascade: Confidence Calibration for Improving the Accuracy and Computational Cost of Cascade Inference Systems}
\author{
    Shohei Enomoto, Takeharu Eda}
\affiliations{

    NTT Software Innovation Center\\

    \{shohei.enomoto.ab, takeharu.eda.bx\}@hco.ntt.co.jp

}

\begin{document}

\maketitle

\begin{abstract}
Recently, deep neural networks have become to be used in a variety of applications.
While the accuracy of deep neural networks is increasing, the confidence score, which indicates the reliability of the prediction results, is becoming more important.
Deep neural networks are seen as highly accurate but known to be overconfident, making it important to calibrate the confidence score.
Many studies have been conducted on confidence calibration.
They calibrate the confidence score of the model to match its accuracy, but it is not clear whether these confidence scores can improve the performance of systems that use confidence scores.
This paper focuses on cascade inference systems, one kind of systems using confidence scores, and discusses the desired confidence score to improve system performance in terms of inference accuracy and computational cost.
Based on the discussion, we propose a new confidence calibration method, Learning to Cascade.
Learning to Cascade is a simple but novel method that optimizes the loss term for confidence calibration simultaneously with the original loss term.
Experiments are conducted using two datasets, CIFAR-100 and ImageNet, in two system settings, and show that naive application of existing calibration methods to cascade inference systems sometimes performs worse.
However, Learning to Cascade always achieves a better trade-off between inference accuracy and computational cost.
The simplicity of Learning to Cascade allows it to be easily applied to improve the performance of existing systems.
\end{abstract}

\section{Introduction}
In recent years, deep learning has dramatically developed and achieved state-of-the-art performance in a variety of applications.
While the accuracy of deep learning has improved over the years, the confidence score, which indicates whether the prediction result is correct or not is becoming more and more important.
In critical decision-making systems such as automated driving \cite{levinson2011towards} and medical diagnostics \cite{miotto2016deep}, humans must make the final decisions when deep neural networks are not confident in their predictive results to prevent serious accidents.
Cascade inference systems \cite{NoScope, IDK, AF, CDL, BranchyNet, MSDNet}, achieve highly accurate and low computational cost inference by combining multiple deep learning models; it decides whether to terminate or continue inference based on the confidence score.
Recent deep neural networks are known to be overconfident and require calibration of the confidence score when applied to these critical systems \cite{TS}.

Several studies have been conducted on confidence calibration \cite{TS, SD, MIXUP, ConfNet,DC}.
These studies have focused on decision-making systems, in which humans decide whether or not to perform inference based on the confidence score of the deep learning model.
It is assumed that humans are almost 100\% accurate.
If the prediction of the model is correct, the prediction result is employed.
And if it is incorrect, humans perform manual inference.
In this way, the system can perform inferences with 100\% accuracy with a small number of human inferences.
To achieve such a best trade-off between accuracy and human effort, it is important to obtain a confidence score that can accurately indicate whether the prediction result of the model is correct or not.

To the best of our knowledge, no study has focused on confidence calibration for cascade inference systems.
Cascade inference is an important technique for performing real-time and accurate inference given limited computing resources such as MEC servers.
It combines more than two models to perform inference: a highly-accurate but expensive model with a low-accuracy but fast model, and determines whether the expensive model should make a prediction or not based on the confidence score of the fast model.
Unlike humans, expensive models are less accurate than 100\%.
If the prediction of the fast model is correct, its result is employed.
If it is incorrect, the expensive model is used instead to make a prediction, but the expensive model prediction may also be incorrect.
At this time, the accuracy of cascade inference does not change, but the computational cost is increased due to the extra calculations incurred by the expensive model.
Therefore, it is important to have a confidence score that reflects not only the prediction result reliability of the fast model but also the prediction result of the expensive model.

In this paper, we discuss confidence scores suitable for cascade inference.
We propose {\it Learning to Cascade (LtC)}, a confidence calibration method for cascade inference.
LtC introduces a new loss term to output confidence scores suitable for cascade inference; it simultaneously optimizes this loss term and the original loss term.
LtC simply adds just a single new loss term and it does not require any changes to the network architecture or optimization method used.
Thus it can be easily added to any existing implementations.

We evaluate Learning to Cascade using CIFAR-100 \cite{CIFAR} and ImageNet \cite{ImageNet}, a standard image classification task.
The effectiveness of LtC is shown by experiments in two settings: model cascading and model splitting.
The experiments show that existing confidence calibration methods may not improve the trade-off between accuracy and the computational cost of inference.
LtC, however, always achieves a better trade-off.

The main contributions of this paper are as follows.
\begin{itemize}
 \item We discuss confidence scores suitable for cascade inference and introduce a new loss term for confidence calibration.
 \item We propose a simple calibration method, LtC, which simultaneously optimizes the original loss term and the loss term for cascade inference. LtC is easily applied to existing implementations because it adds just the new loss term.
 \item In experiments with various settings, we show that LtC achieves the better trade-off between inference accuracy and computational cost than the existing methods.
\end{itemize}

The rest of the paper is organized as follows: 
Section 2 briefly reviews related works in the field of confidence calibration and cascade inference systems. 
The problem statement of cascade inference systems is given in Section 3.
The proposed method, called Learning to Cascade, is described in Section 4. 
The experiments are provided in Section 5. 
Conclusions and future work are summarized in Section 6.

\section{Related Works}

\subsection{Confidence Calibration}
Guo et al. \cite{TS} point out that recent deep learning models are overconfident.
Starting with this paper, several confidence calibration methods are proposed \cite{SD, MIXUP, ConfNet,DC}.
These studies conduct evaluations using some metrics such as Expected Calibration Error (ECE).
As these metrics are calculated solely from the prediction results and labels, it does not evaluate the performance of systems that use confidence scores.
It is not clear whether these calibration methods can improve the performance of systems such as a decision-making system and cascade inference systems.

Individually optimized models may not be optimal for systems that use confidence scores.
Bansal et al. \cite{AI-human} propose a method for training models to optimize the performance of the entire decision-making system.
They assume 100\% human accuracy and evaluate only simple machine learning models and binary classification tasks.
In practice, human accuracy is less than 100\% for the multi-classification tasks commonly used in deep learning (e.g., 94\% for CIFAR-10 \cite{CIFAR}).

Several confidence calibration methods for out-of-distribution (OoD) detection have also been studied \cite{ODIN, GODIN, robustOOD, mahalanobis, CE, GANOOD}.
These studies calibrate confidence scores to detect OoD data.
Although the problem setting is different from this paper, the perspective of confidence calibration to improve the performance of the task (or system) is the same.

\subsection{Cascade Inference Systems}
In this paper, we refer to a system that combines multiple models with different computational costs and accuracies to perform inference as {\it cascade inference systems}.
Cascade inference determines whether the expensive model will make accurate predictions or not based on the confidence score of the fast model.
Cascade inference systems work well with edge-cloud environment and there is a lot of studies in this field \cite{VLDB2018,EI,HOTNETS2018,bottlenet++,bottlenet2}.
There are two types of cascade inference systems, {\it model cascading} and {\it model splitting}.
In model cascading, models with different architectures (e.g., MobileNet \cite{MobileNet, MobileNetV2, MobileNetV3} and ResNet \cite{ResNet}) are combined.
Model splitting allows for dynamic inference by creating a model with multiple early exit points.

Model cascading is originally proposed in the context of face detection in image processing \cite{viola}; recent studies attempt to apply it to speed up heavy computation of deep learning inference.
NoScope is a system that enables fast and accurate object detection as it combines {\it Difference detector}, {\it Specialized Model} and {\it Reference NN} \cite{NoScope}.
It decides whether the Reference NN should make the prediction or not based on the maximum prediction probability of the Specialized Model.
There is another approach that provides an additional model to determine whether to use the fast model or the expensive model to make predictions \cite{IDK, AF, RL}.
The method of preparing additional models can provide accurate confidence scores, but deploying the models increases the computational cost and requires more computational resources.
It also needs extra training effort, such as hyper-parameter tuning.

Instead of connecting different models, the other approaches split a single large model by creating early exit points \cite{CDL, BranchyNet, MSDNet}.
These architectures reduce the computational cost by making most of the easy samples predictions at the earliest possible exit points.
Although harder samples predictions are made at the later exit points, the intermediate layers are shared with the earlier exit points and their computation are completed, so the increase in computational cost is less.
Either the maximum class probability or entropy of the immediately prior exit point is used as the confidence score.
However, these studies focus on just the network architecture and not on the confidence score.

\section{Problem Statement of Cascade Inference}

\begin{figure}[t]
\centering
\includegraphics[width=\linewidth]{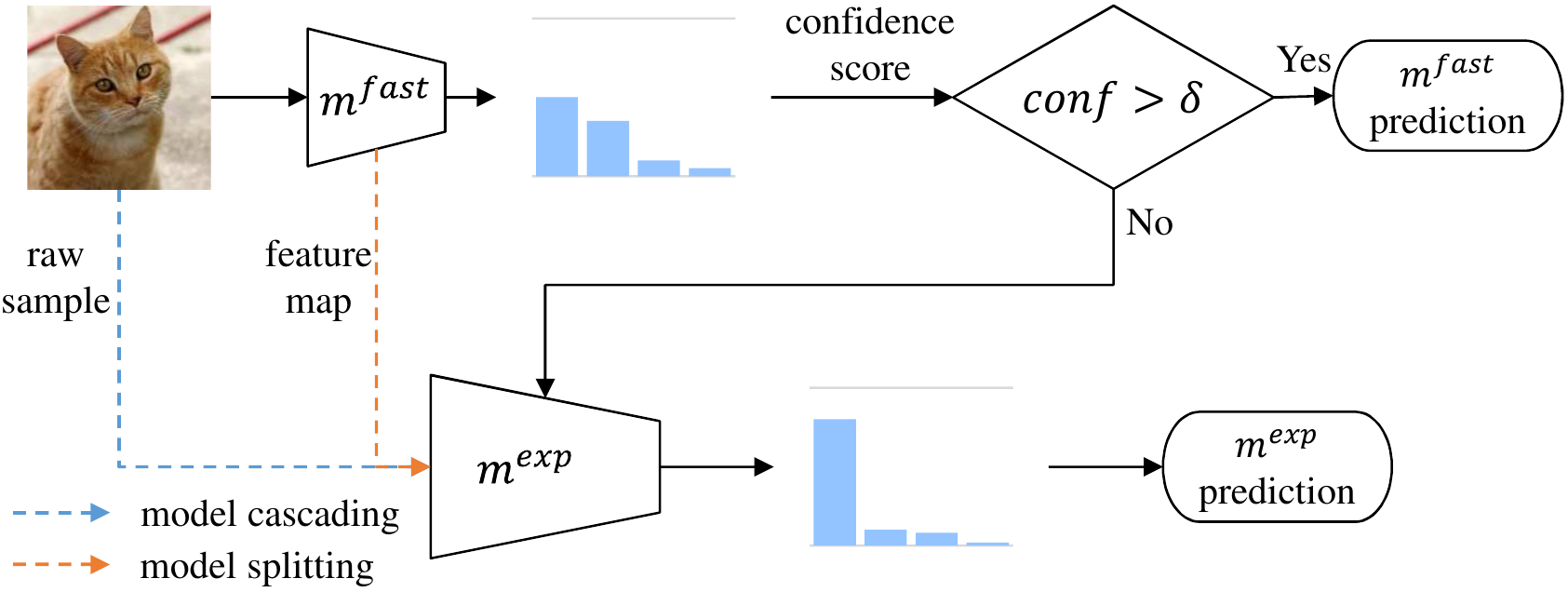}
\caption{Cascade inference systems diagram for two elements case. 
$m^{fast}$ is the fast model, 
$m^{exp}$ is the expensive model.}
\label{fig:model_cascading}
\end{figure}

Figure \ref{fig:model_cascading} shows the system diagram of cascade inference systems for two elements case.
In the model cascading setting, it uses models with different accuracy and computational costs.
In the model splitting setting, each early exit point is treated as a fast or expensive model.
There can be more than two models.
First, the fast model makes a prediction which then confidence score $conf$ is assigned to.
It is easy to use the maximum predicted probability or entropy as the confidence score.
Next, $conf$ is compared with predetermined threshold $\delta$.
If $conf$ is above $\delta$, the prediction result is likely to be correct and the inference is terminated with that result.
If it is below $\delta$, the prediction result is likely to be incorrect, so the expensive model is activated to make a prediction and its output is used as the prediction result.
The inputs of the expensive model are raw samples  and feature maps in the model cascading setting and the model splitting setting, respectively.

This configuration works well with edge-cloud environment.
The fast model is deployed on the edge device as it has limited computing resources, while the expensive model is deployed on the resource-rich cloud. 
Cascade inference can perform most of the inferences using only the fast model on the edge device without losing accuracy. 
This reduces communication costs for data transfer to the cloud and lowers latency due to proximity of the edge device to the application.

The goal of cascade inference is to achieve a good trade-off between accuracy and computational cost of inference.
The accuracy is likely to increase if the expensive model is used, but the computational cost is high and the inference speed is slow.
We begin with dataset $D=\{ (x_i, y_i ) \}^N_{i=1}$, where $x$ are the samples, and $y$ are the category labels for $K$ classes.
The number of samples predicted by the expensive model is given by the following equation.
\begin{eqnarray}
  N^{exp} = \sum^N_{i=1} {\bf 1}_{conf_i \leq \delta}
\end{eqnarray}
${\bf 1}$ is an indicator function that returns $1$ if the content is true and $0$ if it is false.

The problem of cascade inference can be formulated as minimizing $N^{exp}$ while achieving the desired accuracy.
In this paper, the desired accuracy is assumed to be that of the expensive model ($Acc^{exp}$).
It can be formulated as an optimization problem as follows.
\begin{eqnarray}
minimize&&N^{exp} \nonumber \\
subject \: to&&Acc^{casc} \geq (1-\epsilon)Acc^{exp} \nonumber
\end{eqnarray}
$\epsilon$ is an acceptable accuracy degradation and is set to suit the application requirements.
In this paper, we set $\epsilon=0$.
$Acc^{casc}$ is the accuracy by cascade inference and is formulated as the following equation.
\begin{eqnarray}
& Acc^{casc} = \frac{1}{N} \sum^N_{i=1} \{ {\bf 1}_{conf_{i} > \delta} {\bf 1}_{y_i = \mathop{\rm arg~max}\limits_{j} \hat{y}^{fast^{(j)}}_i} \nonumber \\
& \mbox{} + {\bf 1}_{conf_{i} \leq \delta} {\bf 1}_{y_i = \mathop{\rm arg~max}\limits_{j} \hat{y}^{exp^{(j)}}_i} \}
\label{eq:Acc^casc}
\end{eqnarray}
$j$ is the class index, $j \in \{ 1, 2, ..., K \}$.
$\hat{y}^{fast^{(j)}}$ and $\hat{y}^{exp^{(j)}}$ are the predicted probabilities of the fast and expensive models, respectively.
It is possible to obtain the desired trade-off by adjusting $\delta$.
When $\delta$ is close to $0$, the expensive model makes fewer predictions, which decreases the computational cost but accuracy is not expected to increase.
Conversely, when $\delta$ is close to $1$, the expensive model makes more predictions, which increases the computational cost but accuracy is expected to increase.

\section{Learning to Cascade}
\subsection{Confidence Calibration for Cascade Inference}
The existing methods calibrate the confidence score of the model to match its accuracy.
However, it is not clear whether such calibration improves the performance of systems that use confidence scores, such as decision-making systems and cascade inference systems.

In this section, we call the first deep learning model that makes predictions a ``model'' and human or highly-accurate deep learning model that makes decisions based on the confidence scores of its predictions a ``final decision-maker".
Let us assume that the final decision-maker of the systems that use confidence scores is 100\% accurate.
If the systems can accurately determine whether the predictions of the model are correct or not, the accuracy of the systems will be improved.
Because if the prediction result of the model is correct, it is employed; if it is not, the final decision-maker makes the prediction instead and get the correct answer, thus achieving 100\% accuracy with less effort.
We consider that existing confidence calibration methods can improve the performance of these systems because they obtain a confidence score that indicates whether the prediction result of the model is correct or not.
However, in general, neither humans nor highly-accurate deep neural networks will ever be 100\% accurate.
In other words, the final decision-maker can get predictions wrong.
If the predictions of the model and the predictions of the final decision-maker are both incorrect, there is no benefit from the predictions of the final decision-maker.
Therefore, confidence scores should reflect information about whether the final decision-maker is correct or not.
In this paper, we focus on the cascade inference systems problems stated in Section 3 and discuss a confidence score suitable for such systems.

In cascade inference, the predictions of expensive models are costly compared to fast models predictions.
This cost includes computation time, computational resources, and data transfer rate (when in edge-cloud setting), and so on.
If the prediction of the fast model is incorrect but the prediction of the expensive model is correct, then the expensive model should be used because while it is more costly, it can increase accuracy.
However, in other cases, the expensive model prediction will only increase the cost and not the accuracy, even if the expensive model is correct.
Of particular note, in a case in which the prediction of the fast model is correct but the prediction of the expensive model is incorrect, using the expensive model will increase the computational cost and decrease the accuracy.
Therefore, if only the prediction of the expensive model is correct, the confidence score should be smaller.
In other cases, the confidence score should be larger.
Especially, when only the prediction of the fast model is correct, it is most important to obtain a larger confidence score.

Based on the previous discussion, the new loss term for cascade inference is given by the following equation.
\begin{eqnarray}
&L_{casc} =  \frac{1}{N} \sum^N_{i=1} \{ {conf}_{i} {\bf 1}_{y_{i} \neq \mathop{\rm arg~max}\limits_{j} \hat{y}^{fast^{(j)}}_i} \nonumber \\
&\mbox{} + (1 - {conf}_{i})({\bf 1}_{y_{i} \neq \mathop{\rm arg~max}\limits_{j} \hat{y}^{exp^{(j)}}_i} + C) \}
\label{eq:L_casc}
\end{eqnarray}
${conf}$ is the confidence score, which is the maximum softmax prediction probability of the fast model.
$C$ is the parameter accounting for the cost of the expensive model inference.  
It is determined based on system requirements such as computation time, computational resources, and data transfer rate.
In our experiments, we use $C=0.5$.

\subsection{Overall Loss Function}
\begin{figure}[t]
\centering
\includegraphics[width=0.9\linewidth]{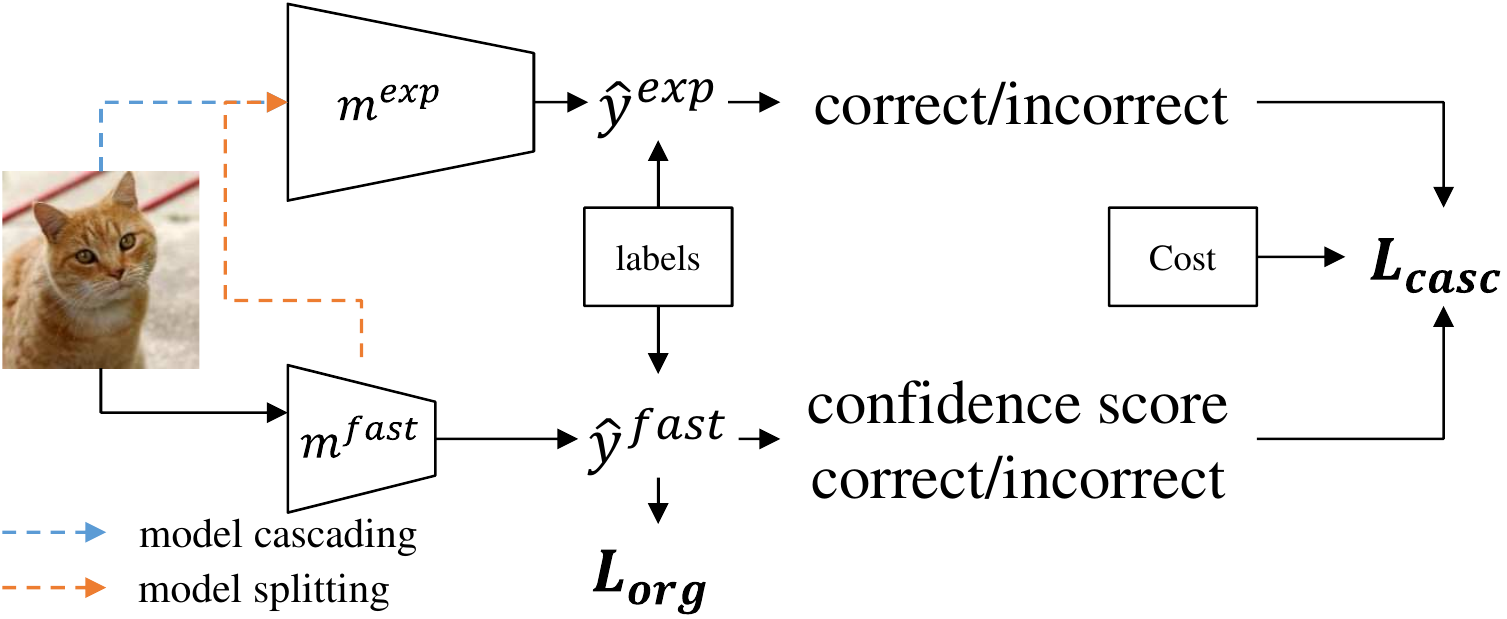}
\caption{Learning to Cascade overview.
$L_{org}$ is calculated from $\hat{y}^{fast}$ and labels. 
$L_{casc}$ is calculated from correct/incorrect of the model prediction result, the value of the confidence score, and cost $C$.}
\label{fig:LtC}
\end{figure}

Using Equation \ref{eq:L_casc}, we propose a confidence calibration method for cascade inference, called Learning to Cascade (LtC).
LtC simultaneously optimizes the original loss term and the loss term for cascade inference; it improves classification accuracy and calibrates the confidence score.
In this paper, we use softmax cross entropy as the original loss term, $L_{org}$.
The overall loss function is as follows.
\begin{eqnarray}
L =L_{org} + wL_{casc}
\end{eqnarray}
$w$ is a parameter that determines the weight of the loss term for cascade inference.
An overview of LtC is shown in Figure \ref{fig:LtC}.

\subsection{Beyond the Two Elements Cascade} 
LtC can be applied to cascade inference systems even when there are $M$ models ($M \geq 3$).
The $M$ models are sorted in order of inference speed and $model^m$ is the $m$-th model ($m \in \{ 1, 2, ..., M \}$).

In the model cascading setting, $model^M$ is initially trained by the original loss function.
Next, $model^{M-1}$ is trained by the LtC loss function with $model^M$ as the expensive model.
This procedure is repeated until all models are trained.
The loss function of $model^m$ is given by the following equation.
\begin{eqnarray}
L^{(m)} = \left\{ \begin{array}{ll}
L_{org}^{(M)} & (m = M) \\
L_{org}^{(m)} + wL_{casc}^{(m, m+1)} & (m \leq M-1) \\
\end{array} \right.
\end{eqnarray} 

In the model splitting setting, each early exit classifier is treated as a single model.
Unlike the model cascading setting, all models are trained at the same time.
The $m$-th early exit point is treated as a fast model ($model^m$) and the $(m+1)$-th early exit point as an expensive model ($model^{m+1}$).
LtC computes $L_{casc}$ for all pairs of $model^m$ and $model^{m+1}$ and adds their sum to the original loss term.
The loss function is given by the following equation.
\begin{eqnarray}
&L =  \sum^{M-1}_{m=1} \{ L_{org}^{(m)} + wL_{casc}^{(m, m+1)} \} + L_{org}^{(M)}
\label{eq:a}
\end{eqnarray}

\section{Experiments}
We conducted experiments on two datasets and in two settings to show that LtC achieves a good trade-off between accuracy and computational cost.
\subsection{Experimental Setup}
\subsubsection{Datasets:}
We used CIFAR-100 \cite{CIFAR} and ImageNet (ILSVRC 2012) \cite{ImageNet}.
CIFAR-100 is composed of 50,000 training images and 10,000 test images.
We randomly split the training images into training and validation images at the ratio of  9:1.
ImageNet is composed of 1.3 million training images and 50,000 test images.
We randomly split the training images into training and validation images at the ratio of  99:1.
The best values of parameters of LtC, $\delta$ and $w$ were searched using the validation images.
The CIFAR-100 experiment was conducted 5 times with different random seeds, and the ImageNet experiment was conducted 3 times with different random seeds.

\subsubsection{Networks and Training Details:}
The experiments were conducted in two settings, model cascading and model splitting.

In the model cascading setting, we used AlexNet \cite{AlexNet}, VGG11 \cite{VGG} and MobileNetV2 \cite{MobileNetV2} for the fast model, and ResNet18 and ResNet152 \cite{ResNet} for the expensive model.
We used SGD with momentum $0.9$ as the optimizer.
In CIFAR-100 experiments, all models were trained with batch size of $128$ for $200$ epochs with weight decay of $0.0005$. 
The learning rate started  with $0.1$ and decreased by the factor of $0.2$ at $60$, $120$, $160$ training epochs.
In ImageNet experiments, all models were trained with batch size of $256$ for $100$ epochs with weight decay of $0.0001$. 
The learning rate started  with $0.1$ and decreased by the factor of $0.1$ at $30$, $60$, $90$ training epochs.
Table \ref{table:model_acc_MACs} shows the accuracy and MACs of each model.
\begin{table}[t]
\begin{center}
\small
\begin{tabular}{cccc}
\hline
Dataset & Model & Accuracy{[}\%{]} &  MACs \\ \hline
CIFAR-100 & AlexNet & $ 65.57 \pm 0.16$ & 285.4[M] \\
 & VGG11 & $ 67.26 \pm 0.17$ & 191.8[M] \\
 & MobileNetV2 & $67.18 \pm 0.10$ & 67.6[M] \\
 & ResNet18 & $74.63 \pm 0.08$ & 556.8[M] \\ 
 & ResNet152 & $77.20 \pm 0.28$ & 3737[M] \\ \hline
ImageNet & MobileNetV2 & $68.52 \pm 0.12$ & 314.13[M] \\
 &  ResNet152 & $77.24 \pm 0.07$ & 11.559[G] \\ \hline
\end{tabular}
\caption{Accuracy and MACs of evaluated models results.}
\label{table:model_acc_MACs}
\end{center}
\end{table}

In the model splitting setting, we used MSDNet \cite{MSDNet}, the state-of-the-art early exit architecture for image classification tasks.
MSDNet has several parameters that determine the architecture.
$nBlocks$ is the number of early exit points, $base$ is the number of layers in the first block, and $step$ is the number of layers after the second block.
We tested several combinations of parameters.
We also used a setting wherein the number of layers after the second block increased linearly.
MSDNet was trained in the same configuration as \cite{MSDNet}.

\subsubsection{Evaluation Metrics:}
Two metrics were used to evaluate the results.
The first is $Acc^{casc}$, which is the accuracy in cascade inference shown in Equation \ref{eq:Acc^casc}.
The second is multiply-accumulate ($MACs^{casc}$), which is the computational cost of cascade inference given by the following equation.
\begin{eqnarray}
 {MACs}^{casc} = {MACs}^{fast} + \frac{N^{exp}}{N} {MACs}^{exp}
\label{eq:MACs}
\end{eqnarray}
${MACs}^{fast}$ and ${MACs}^{exp}$ are MACs for the fast and expensive models, respectively.
Based on the formulation in Section 3, $Acc^{casc}$ should be larger than or equal to $Acc^{exp}$; the smaller the $MACs^{casc}$ is, the better the system is.

\subsubsection{Methods for Comparison:}
In the model cascading setting, we compared LtC with four existing methods.
\begin{itemize}
 \item {\bf Baseline}: Baseline is a method using the maximum predicted probability of the fast model as the confidence score.
 \item {\bf IDK}: IDK Cascades \cite{IDK} employs an additional model to obtain confidence scores for the model cascading.
We used a CNN with one fully connected layer for an additional model (Same model as ConfNet \cite{ConfNet}). 
 \item {\bf ConfNet}: ConfNet \cite{ConfNet} also employs an additional model to obtain accurate confidence scores but does not consider cascade inference.
We used a CNN with one fully connected layer as the additional model.
 \item {\bf Temp. Scaling}: Temperature Scaling \cite{TS} is the general confidence calibration method attained by dividing the logits with the temperature $T$.
\end{itemize}

\subsection{Results}
\subsubsection{The Model Cascading Setting:}
Model cascading performs inferences by combining multiple models with different computational costs and accuracies.
We used a set of validation images to search for the $\delta$ with the highest cascade accuracy and used that value for testing.

In the CIFAR-100 experiment, we first examined two elements cascade.
Three different fast models and two different expensive models were used.
The accuracy and MACs results for model cascading setting are shown in Table \ref{tab:model_cascade_acc}, \ref{tab:model_cascade_MACs}.
\begin{table*}[t]
\begin{center}
\small
\begin{tabular}{ccc}
\hline
$m^{fast}$ & $m^{exp}$ & $Acc^{casc}${[}\%{]} \\ \hline
 &  & Baseline / IDK / ConfNet / Temp. Scaling / LtC \\ \hline
AlexNet       &               & $74.60 \pm 0.08$ /  $74.55 \pm 0.10$ / $74.60 \pm 0.09$ / ${\bf 74.62 \pm 0.07}$ / $74.60 \pm 0.07$  \\ 
VGG11        & ResNet18 & ${\bf 74.69 \pm 0.07}$ / $74.64 \pm 0.10$ / $74.60 \pm 0.07$ / ${\bf 74.69 \pm 0.09}$ / $74.66 \pm 0.09$  \\ 
MobileNetV2 &               & $74.71 \pm 0.09$ / $74.69 \pm 0.06$ / ${\bf 74.72 \pm 0.09}$ / $74.71 \pm 0.09$ / $74.69 \pm 0.08$   \\ \hline
AlexNet       &          & $77.17 \pm 0.30$ / ${\bf 77.18 \pm 0.30}$ / ${\bf 77.18 \pm 0.28}$ / $77.17 \pm 0.28$ /  $77.02 \pm 0.28$  \\ 
VGG11 & ResNet152 & ${\bf 77.19 \pm 0.28}$ / $77.13 \pm 0.28$ / $77.18 \pm 0.27$ / ${\bf 77.19 \pm 0.27}$ / $77.12 \pm 0.30$   \\ 
MobileNetV2  &         & ${\bf 77.22 \pm 0.28}$ / ${\bf 77.22 \pm 0.28}$ / $77.21 \pm 0.28$ / $77.20 \pm 0.28$ / $77.15 \pm 0.26$  \\ \hline 
\end{tabular}
\caption{Accuracy comparison results of model cascading setting on CIFAR-100.}
\label{tab:model_cascade_acc}
\end{center}
\end{table*}
\begin{table*}[t]
\begin{center}
\small
\begin{tabular}{ccc}
\hline
$m^{fast}$ & $m^{exp}$ &  $MACs^{casc}${[}M{]} \\ \hline
 &  & Baseline / IDK / ConfNet / Temp. Scaling / LtC \\ \hline
AlexNet             &  & $722.56 \pm 50.70$ / $650.78 \pm 24.70$ / $841.47 \pm 0.46$ / $678.22 \pm 14.21$ / ${\bf 617.88 \pm 8.92}$ \\ 
VGG11 & ResNet18 &  $527.92 \pm 11.40$ / ${\bf 484.99 \pm 9.00}$ / $747.87 \pm 0.22$ / $562.64 \pm 13.05$ / $ 485.83 \pm 11.15$ \\ 
MobileNetV2        & &  $435.91 \pm 23.80$ / $424.07 \pm 29.56$ / $421.17 \pm 10.18$ / $450.95 \pm 18.09$ / ${\bf 383.08 \pm 13.85}$ \\ \hline
AlexNet               &  &   $3748.18 \pm 274.22$ / $3451.98 \pm 275.01$ / $4020.01 \pm 0.63$ / $ 3141.66 \pm 74.29$ / ${\bf 2720.73 \pm 64.24}$ \\ 
VGG11 & ResNet152 &   $3089.84 \pm 344.71$ /  $2411.88 \pm 59.33$ /  $3925.21 \pm 1.20$ /  $2819.06 \pm 99.87$ / ${\bf 2381.08 \pm 66.62}$ \\ 
MobileNetV2         &  &  $3100.55 \pm 238.31$ /  $2990.83 \pm 233.00$ /  $3122.52 \pm 282.78$ /  $3019.76 \pm 225.20$ / ${\bf 2645.08 \pm 136.97}$ \\ \hline 
\end{tabular}
\caption{MACs comparison results of model cascading setting on CIFAR-100.}
\label{tab:model_cascade_MACs}
\end{center}
\end{table*}
From Tables \ref{table:model_acc_MACs} and \ref{tab:model_cascade_acc},  the mean values of accuracy for all methods and the expensive model differed slightly, but were within the standard error.
All methods meet the condition ($Acc^{casc} \geq Acc^{exp}$) as defined by Section 3.
From Tables \ref{table:model_acc_MACs} and \ref{tab:model_cascade_MACs}, all methods achieve lower MACs than the expensive model, except for the combination of AlexNet and ResNet18.
Furthermore, LtC achieves the lowest $MACs^{casc}$ in many cases.
LtC reduces the MACs of ResNet18 and ResNet152 by up to 31\% and 36\%, respectively, with no accuracy degradation.
LtC achieves a better trade-off between accuracy and computational cost than existing methods.
The reason why AlexNet does not improve the trade-off is in its classic architecture; it does not have so smaller MACs compared to that of ResNet18.
As equation \ref{eq:MACs} shows, MACs of the fast model needs to be relatively smaller than that of the expensive model in order to achieve a better trade-off in cascade inference systems.
The confidence calibration methods, ConfNet and Temperature Scaling, sometimes yield worse results than Baseline.
As discussed in Section 4, a desirable confidence score in cascade inference systems should reflect not only the prediction of the fast model but also the prediction of the expensive model.
Since these methods calibrate the confidence score to a value that indicates only whether the prediction of the fast model is correct or not, these methods cannot obtain the desirable confidence score.
As a result, these methods may not improve the accuracy and computational cost of cascade inference systems.

Next, we examined three elements cascade.
We used MobileNetV2 as the fast model, ResNet18 as the middle fast model and ResNet152 as the expensive model.
The results are shown in Table \ref{tab:3model_cascade}.
\begin{table}[t]
\begin{center}
\small
\begin{tabular}{cc}
\hline
 $Acc^{casc}${[}\%{]} &  $MACs^{casc}${[}G{]} \\ \hline
 \multicolumn{2}{c}{ Baseline / LtC } \\ \hline
$77.48 \pm 0.27$ / ${\bf 77.53 \pm 0.18}$ & $1.91 \pm 0.22$ / ${\bf 1.68 \pm 0.13}$  \\ \hline 
\end{tabular}
\caption{Accuracy and MACs comparison results of three elements cascade on CIFAR-100.}
\label{tab:3model_cascade}
\end{center}
\end{table}
By using LtC, cascade inference can reduce the computational cost by 55\% while improving the accuracy of ResNet152.

In the ImageNet experiment, we used MobileNetV2 as the fast model and ResNet152 as the expensive model.
The accuracy and MACs results for model cascading setting are shown in Table \ref{tab:model_cascade_in}.
\begin{table}[t]
\begin{center}
\small
\begin{tabular}{cc}
\hline
$Acc^{casc}${[}\%{]} &  $MACs^{casc}${[}G{]} \\ \hline
\multicolumn{2}{c}{ Baseline / LtC } \\ \hline
$77.22 \pm 0.10$ / ${\bf 77.24 \pm 0.08}$ & $8.36 \pm 0.65$ / ${\bf 8.08 \pm 0.21}$  \\ \hline 
\end{tabular}
\caption{Accuracy and MACs comparison results of model cascading setting on ImageNet.}
\label{tab:model_cascade_in}
\end{center}
\end{table}
LtC reduces the  MACs of ResNet152 by about  30\% , with no accuracy degradation.
LtC also achieves a better trade-off between accuracy and computational cost in difficult tasks such as ImageNet.

\subsubsection{The Model Splitting Setting:}
\begin{figure}[tb]
	\begin{center}
		\includegraphics[width=\linewidth]{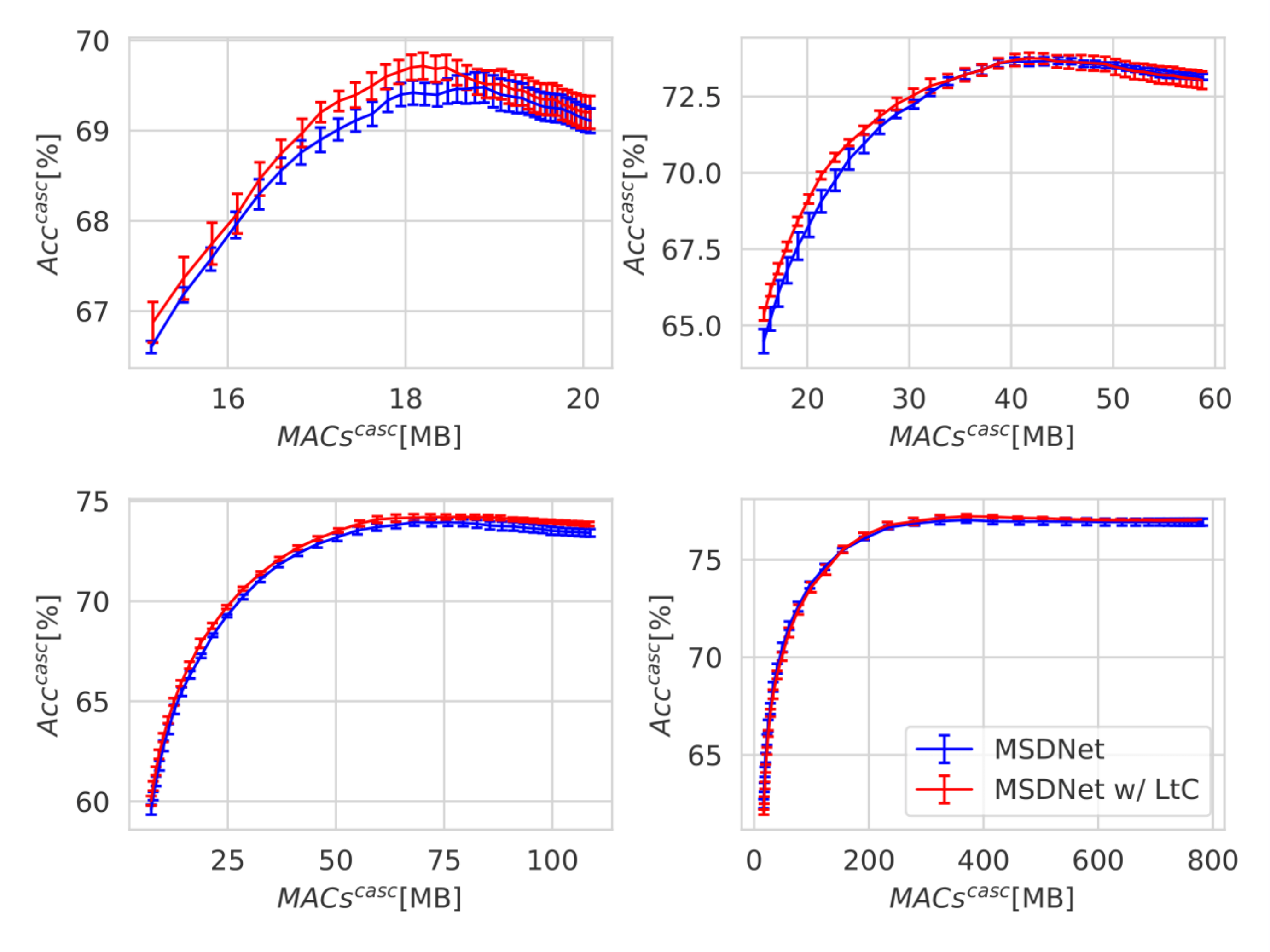}
		\caption{MSDNet and MSDNet w/ LtC comparison results. MSDNet parameters are 
nBlocks=2, step=2, base=4 (top left), 
nBlocks=5, step=1, base=4 (top right),  
nBlocks=7, step=1, base=1 (bottom left),  
nBlocks=10, step=2, base=4 (bottom right), respectively.}
		\label{fig:MSDNet}
	\end{center}
\end{figure}

We evaluated LtC using MSDNet with several network architecture parameters.
Experiments were conducted using CIFAR-100 only.
The results are shown in Figure \ref{fig:MSDNet}. 
For all parameter settings, LtC achieves a better trade-off between accuracy and computational cost than the original MSDNet.
Since we only need to add just a loss term, it can be easily applied to state-of-the-art architectures such as MSDNet.
LtC shows its simplicity and effectiveness by this result.

\subsection{Discussion}
\subsubsection{Effect of LtC on the accuracy of the fast model:}
Since LtC simultaneously optimizes the loss term of the original task and the loss term for cascade inference, we investigated the effect of LtC on the original classification task.
Table \ref{table:model_acc} shows the accuracy of each model trained by only the original loss function and by the LtC loss function.
\begin{table}[tb]
\begin{center}
\small
\begin{tabular}{ccc}
\hline
Dataset & Model & Difference  \\ \hline
CIFAR-100 & AlexNet(w/ ResNet18 LtC) & $-0.30$ \\
& AlexNet(w/ ResNet152 LtC) & $-0.48$  \\ 
 & VGG11(w/ ResNet18 LtC) & $-0.21$  \\
 & VGG11(w/ ResNet152 LtC) & $-0.21$ \\ 
 & MobileNetV2(w/ ResNet18 LtC) & $ +0.10$ \\
 & MobileNetV2(w/ ResNet152 LtC) & $+0.08$ \\ \hline
ImageNet & MobileNetV2(w/ ResNet152 LtC) & $+0.28$  \\ \hline
\end{tabular}
\caption{The difference in accuracy between the model trained by the LtC loss function and the model trained by the original loss function. 
The `w/' refers to which model was used as the expensive model.}
\label{table:model_acc}
\end{center}
\end{table}
Note that this is not the accuracy of the cascade inference, but the accuracy of the fast model.
The model accuracy trained by the LtC loss function is a little lower in AlexNet and VGG11, but a little higher in MobileNetV2.
We consider that adding the loss term for cascade inference does not adversely affect the classification accuracy of the fast model.

\subsubsection{Effects of LtC Parameters:}
\begin{figure}[tb]
	\begin{center}
		\includegraphics[width=\linewidth]{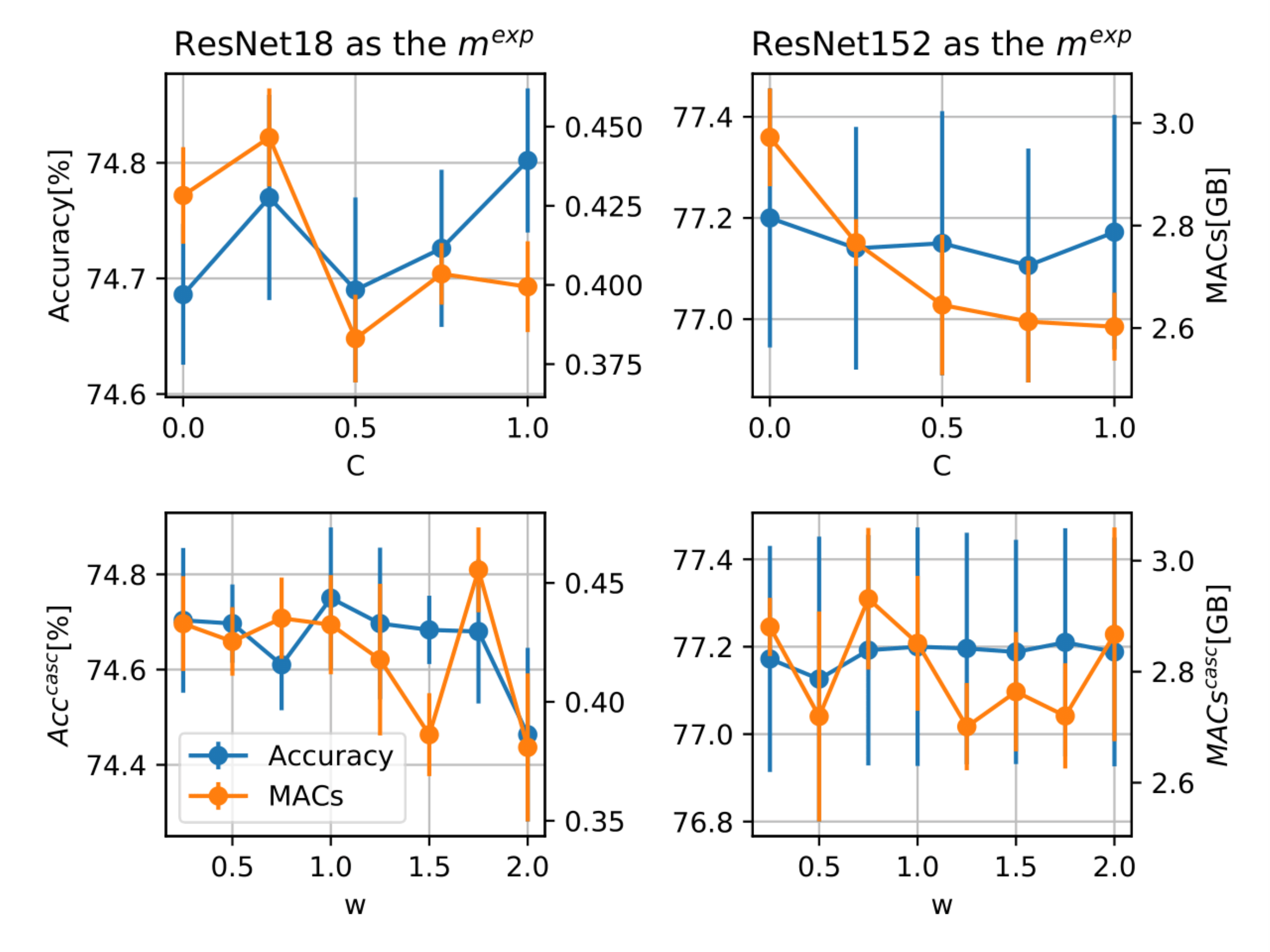}
		\caption{The relationship between LtC parameters ($C$ and $w$) and inference performance ($Acc^{casc}$ and $MACs^{casc}$).
The expensive model is set to ResNet18 (left), 
the expensive model is set to ResNet152 (right), 
when $C$ is varied (top) 
and when $w$ is varied (bottom).}
		\label{fig:LtC_params}
	\end{center}
\end{figure}

LtC has two parameters, $C$ and $w$.
We investigated the relationship between the two parameters and inference performance(Accuracy, MACs).
We examined the model cascading setting using MobileNetV2 as the fast model and ResNet18 and ResNet152 as the expensive model.
The results are shown in Figure \ref{fig:LtC_params}. 
To summarize, $C$ is not correlated with $Acc^{casc}$, but it is correlated with $MACs^{casc}$.
$C$ is the prediction cost of the expensive model.
When $C$ is large, the expensive model makes fewer predictions; when $C$ is small, the expensive model makes more predictions.
$C$ and $MACs^{casc}$ are inversely proportional to each other.
$C$ is a parameter that should be determined based on the system requirements, such as inference speed or computational resources available.
On the other hand, there is no correlation between $w$ and inference performance.
In order to obtain a good trade-off in cascade inference system, we need to search for the optimal $w$.

\subsubsection{Analysis of How LtC Works:}
\begin{figure}[tb]
	\begin{center}
		\includegraphics[width=\linewidth]{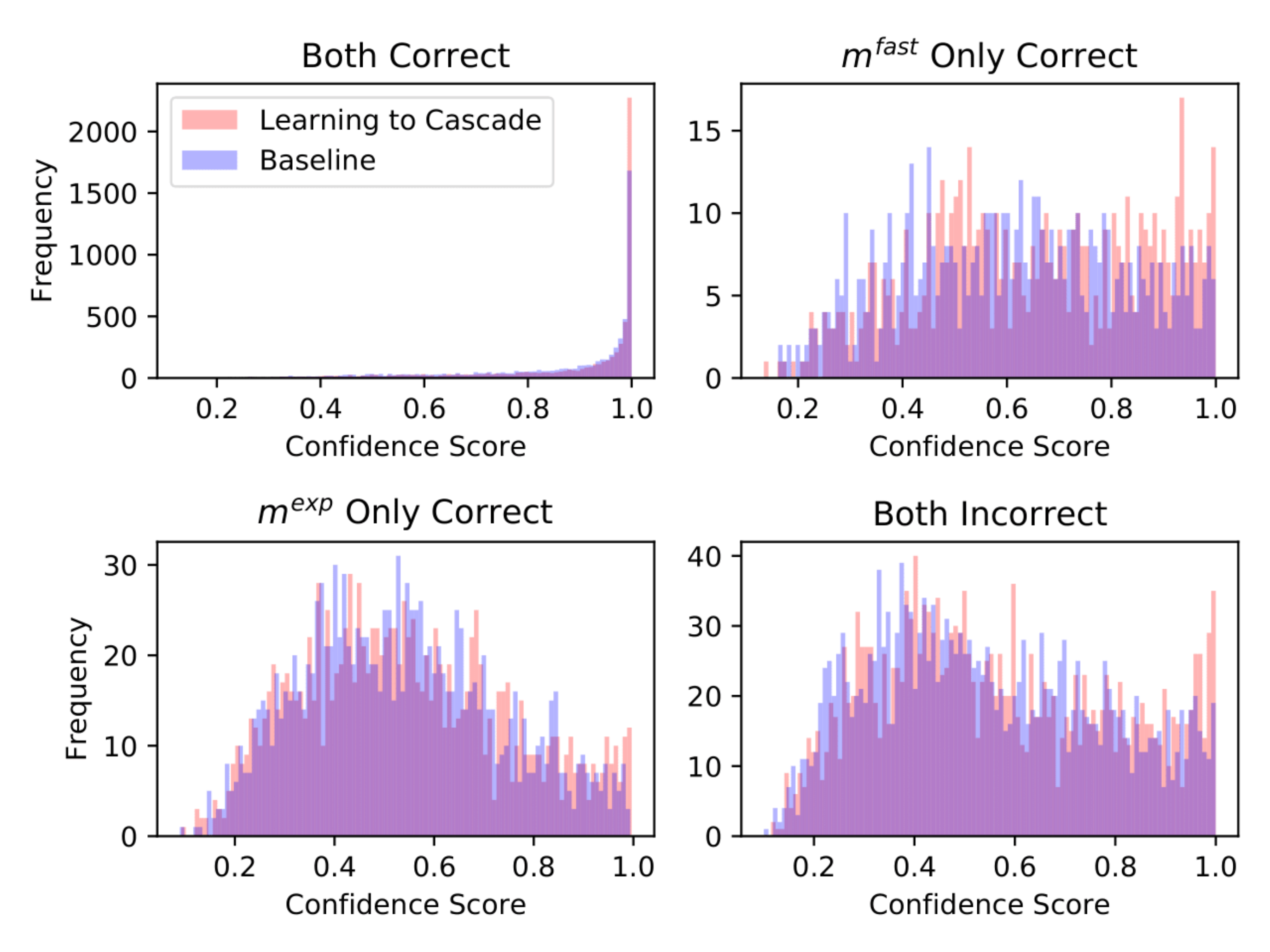}
		\caption{Confidence comparison of LtC and Baseline.
Both prediction results are correct (top left), 
only the $m^{fast}$ prediction results are correct (top right), 
only the $m^{exp}$ prediction results are correct (bottom left), 
and both prediction results are incorrect (bottom right).}
		\label{fig:conf}
	\end{center}
\end{figure}

Given combination of MobileNetV2 and ResNet18, we compared confidence scores of Baseline and LtC for each combination of the fast model and the expensive model prediction result.
The results are shown in Figure \ref{fig:conf}.
In all combinations, LtC has high confidence scores.
The number of confidence scores is especially high around 1.
According to the discussion in Section 4, a low confidence score is desirable for case "$m^{exp}$ Only Correct" and a high confidence score is desirable for the other cases.
So, LtC has a negative impact on case "$m^{exp}$ Only Correct", but a positive impact on the other cases.
In particular, LtC obtains a high confidence score in case "$m^{fast}$ Only Correct", which is the most important case because of its impact on both accuracy and computational cost.
We consider that LtC can improve the performance of cascade inference because its good effects outweighed its negative impact.

\section{Conclusion}
In this paper, we propose Learning to Cascade (LtC), a confidence calibration method for cascade inference.
Existing confidence calibration methods implicitly assume that human accuracy is 100\%.
However, neither humans nor deep learning models can achieve 100\% accuracy for difficult tasks such as those targeted by deep learning.
Therefore, we optimize the confidence score not only for the model but also for the entire system, including the decision-maker (human or more accurate deep learning model).
To improve system performance, we introduce a new loss term for confidence calibration.
LtC simultaneously optimizes this loss term and the original loss term.
We evaluate LtC in two datasets and two system settings.
LtC achieves a low computational cost in most cases with almost the same level of accuracy as existing methods and the expensive model.
In addition, in the three elements cascade configuration, LtC achieves higher accuracy than ResNet152 at 55\% lower computational cost.
It is also shown that existing confidence calibration methods fail to improve accuracy and computational cost in some cases.
Because of its simplicity, LtC can be easily applied to state-of-the-art architectures such as MSDNet to improve a trade-off, which is also verified in our experiments.

In future work, we would like to elucidate the cause of the high confidence score in the case of  "$m^{exp}$ Only Correct" and improve the loss term.
We will also show the effectiveness of LtC for other architectures and tasks.

\bibliography{10_egbib.bib}

\begin{thebibliography}{35}
\providecommand{\natexlab}[1]{#1}
\providecommand{\url}[1]{\texttt{#1}}
\providecommand{\urlprefix}{URL }
\expandafter\ifx\csname urlstyle\endcsname\relax
  \providecommand{\doi}[1]{doi:\discretionary{}{}{}#1}\else
  \providecommand{\doi}{doi:\discretionary{}{}{}\begingroup
  \urlstyle{rm}\Url}\fi

\bibitem[{Bansal et~al.(2020)Bansal, Nushi, Kamar, Horvitz, and
  Weld}]{AI-human}
Bansal, G.; Nushi, B.; Kamar, E.; Horvitz, E.; and Weld, D.~S. 2020.
\newblock Optimizing AI for Teamwork.
\newblock \emph{arXiv preprint arXiv:2004.13102} .

\bibitem[{Chen et~al.(2020)Chen, Wu, Liang, Jha et~al.}]{robustOOD}
Chen, J.; Wu, X.; Liang, Y.; Jha, S.; et~al. 2020.
\newblock Robust Out-of-distribution Detection in Neural Networks.
\newblock \emph{arXiv preprint arXiv:2003.09711} .

\bibitem[{Chinchali et~al.(2019)Chinchali, Sharma, Harrison, Elhafsi, Kang,
  Pergament, Cidon, Katti, and Pavone}]{RL}
Chinchali, S.; Sharma, A.; Harrison, J.; Elhafsi, A.; Kang, D.; Pergament, E.;
  Cidon, E.; Katti, S.; and Pavone, M. 2019.
\newblock Network Offloading Policies for Cloud Robotics: a Learning-based
  Approach.
\newblock \emph{arXiv preprint arXiv:1902.05703} .

\bibitem[{Chinchali et~al.(2018)Chinchali, Cidon, Pergament, Chu, and
  Katti}]{HOTNETS2018}
Chinchali, S.~P.; Cidon, E.; Pergament, E.; Chu, T.; and Katti, S. 2018.
\newblock Neural Networks Meet Physical Networks: Distributed Inference Between
  Edge Devices and the Cloud.
\newblock In \emph{Proceedings of the 17th ACM Workshop on Hot Topics in
  Networks}, 50--56.

\bibitem[{DeVries and Taylor(2018)}]{CE}
DeVries, T.; and Taylor, G.~W. 2018.
\newblock Learning Confidence for Out-of-distribution Detection in Neural
  Networks.
\newblock \emph{arXiv preprint arXiv:1802.04865} .

\bibitem[{Grulich and Nawab(2018)}]{VLDB2018}
Grulich, P.~M.; and Nawab, F. 2018.
\newblock Collaborative Edge and Cloud Neural Networks for Real-time Video
  Processing.
\newblock \emph{Proceedings of the VLDB Endowment} 11(12): 2046--2049.

\bibitem[{Guo et~al.(2017)Guo, Pleiss, Sun, and Weinberger}]{TS}
Guo, C.; Pleiss, G.; Sun, Y.; and Weinberger, K.~Q. 2017.
\newblock On Calibration of Modern Neural Networks.
\newblock In \emph{Proceedings of the 34th International Conference on Machine
  Learning}, volume~70, 1321--1330.

\bibitem[{He et~al.(2016)He, Zhang, Ren, and Sun}]{ResNet}
He, K.; Zhang, X.; Ren, S.; and Sun, J. 2016.
\newblock Deep Residual Learning for Image Recognition.
\newblock In \emph{Proceedings of the IEEE Conference on Computer Vision and
  Pattern Recognition}, 770--778.

\bibitem[{Howard et~al.(2019)Howard, Sandler, Chu, Chen, Chen, Tan, Wang, Zhu,
  Pang, Vasudevan et~al.}]{MobileNetV3}
Howard, A.; Sandler, M.; Chu, G.; Chen, L.; Chen, B.; Tan, M.; Wang, W.; Zhu,
  Y.; Pang, R.; Vasudevan, V.; et~al. 2019.
\newblock Searching for Mobilenetv3.
\newblock In \emph{Proceedings of the IEEE International Conference on Computer
  Vision}, 1314--1324.

\bibitem[{Howard et~al.(2017)Howard, Zhu, Chen, Kalenichenko, Wang, Weyand,
  Andreetto, and Adam}]{MobileNet}
Howard, A.~G.; Zhu, M.; Chen, B.; Kalenichenko, D.; Wang, W.; Weyand, T.;
  Andreetto, M.; and Adam, H. 2017.
\newblock Mobilenets: Efficient Convolutional Neural Networks for Mobile Vision
  Applications.
\newblock \emph{arXiv preprint arXiv:1704.04861} .

\bibitem[{Hsu et~al.(2020)Hsu, Shen, Jin, and Kira}]{GODIN}
Hsu, Y.; Shen, Y.; Jin, H.; and Kira, Z. 2020.
\newblock Generalized Odin: Detecting Out-of-distribution Image without
  Learning from Out-of-distribution Data.
\newblock In \emph{Proceedings of the IEEE Conference on Computer Vision and
  Pattern Recognition}, 10951--10960.

\bibitem[{Huang et~al.(2018)Huang, Chen, Li, Wu, van~der Maaten, and
  Weinberger}]{MSDNet}
Huang, G.; Chen, D.; Li, T.; Wu, F.; van~der Maaten, L.; and Weinberger, K.
  2018.
\newblock Multi-Scale Dense Networks for Resource Efficient Image
  Classification.
\newblock In \emph{International Conference on Learning Representations}.

\bibitem[{Kang et~al.(2017)Kang, Emmons, Abuzaid, Bailis, and
  Zaharia}]{NoScope}
Kang, D.; Emmons, J.; Abuzaid, F.; Bailis, P.; and Zaharia, M. 2017.
\newblock NoScope: Optimizing Neural Network Queries Over Video at Scale.
\newblock \emph{Proceedings of the VLDB Endowment} 10(11): 1586--1597.

\bibitem[{Krizhevsky, Hinton et~al.(2009)}]{CIFAR}
Krizhevsky, A.; Hinton, G.; et~al. 2009.
\newblock Learning Multiple Layers of Features from Tiny Images.
\newblock Technical report, Citeseer.

\bibitem[{Krizhevsky, Sutskever, and Hinton(2012)}]{AlexNet}
Krizhevsky, A.; Sutskever, I.; and Hinton, G.~E. 2012.
\newblock Imagenet Classification with Deep Convolutional Neural Networks.
\newblock In \emph{Advances in Neural Information Processing Systems},
  1097--1105.

\bibitem[{Kull et~al.(2019)Kull, Perello~Nieto, K\"{a}ngsepp, Silva~Filho,
  Song, and Flach}]{DC}
Kull, M.; Perello~Nieto, M.; K\"{a}ngsepp, M.; Silva~Filho, T.; Song, H.; and
  Flach, P. 2019.
\newblock Beyond Temperature Scaling: Obtaining Well-calibrated Multi-class
  Probabilities with Dirichlet Calibration.
\newblock In \emph{Advances in Neural Information Processing Systems},
  12316--12326.

\bibitem[{Lee et~al.(2018)Lee, Lee, Lee, and Shin}]{mahalanobis}
Lee, K.; Lee, K.; Lee, H.; and Shin, J. 2018.
\newblock A Simple Unified Framework for Detecting Out-of-distribution Samples
  and Adversarial Attacks.
\newblock In \emph{Advances in Neural Information Processing Systems},
  7167--7177.

\bibitem[{LEE et~al.(2018)LEE, Lee, Lee, and Shin}]{GANOOD}
LEE, K.; Lee, K.; Lee, H.; and Shin, J. 2018.
\newblock Training Confidence-calibrated Classifiers for Detecting
  Out-of-distribution Samples.
\newblock In \emph{International Conference on Learning Representations}.

\bibitem[{Levinson et~al.(2011)Levinson, Askeland, Becker, Dolson, Held,
  Kammel, Kolter, Langer, Pink, Pratt et~al.}]{levinson2011towards}
Levinson, J.; Askeland, J.; Becker, J.; Dolson, J.; Held, D.; Kammel, S.;
  Kolter, J.~Z.; Langer, D.; Pink, O.; Pratt, V.; et~al. 2011.
\newblock Towards Fully Autonomous Driving: Systems and Algorithms.
\newblock In \emph{IEEE Intelligent Vehicles Symposium}, 163--168.

\bibitem[{Li, Zhou, and Chen(2018)}]{EI}
Li, E.; Zhou, Z.; and Chen, X. 2018.
\newblock Edge Intelligence: On-demand Deep Learning Model Co-inference with
  Device-edge Synergy.
\newblock In \emph{Proceedings of the 2018 Workshop on Mobile Edge
  Communications}, 31--36.

\bibitem[{Liang, Li, and Srikant(2018)}]{ODIN}
Liang, S.; Li, Y.; and Srikant, R. 2018.
\newblock Enhancing the Reliability of Out-of-distribution Image Detection in
  Neural Networks.
\newblock In \emph{International Conference on Learning Representations}.

\bibitem[{Miotto et~al.(2016)Miotto, Li, Kidd, and Dudley}]{miotto2016deep}
Miotto, R.; Li, L.; Kidd, B.~A.; and Dudley, J.~T. 2016.
\newblock Deep Patient: An Unsupervised Representation to Predict the Future of
  Patients from the Electronic Health Records.
\newblock \emph{Scientific Reports} 6(1): 1--10.

\bibitem[{Panda, Sengupta, and Roy(2016)}]{CDL}
Panda, P.; Sengupta, A.; and Roy, K. 2016.
\newblock Conditional Deep Learning for Energy-efficient and Enhanced Pattern
  Recognition.
\newblock In \emph{Design, Automation and Test in Europe Conference and
  Exhibition}, 475--480.

\bibitem[{Russakovsky et~al.(2015)Russakovsky, Deng, Su, Krause, Satheesh, Ma,
  Huang, Karpathy, Khosla, Bernstein et~al.}]{ImageNet}
Russakovsky, O.; Deng, J.; Su, H.; Krause, J.; Satheesh, S.; Ma, S.; Huang, Z.;
  Karpathy, A.; Khosla, A.; Bernstein, M.; et~al. 2015.
\newblock Imagenet Large Scale Visual Recognition Challenge.
\newblock \emph{International Journal of Computer Vision} 115(3): 211--252.

\bibitem[{Sandler et~al.(2018)Sandler, Howard, Zhu, Zhmoginov, and
  Chen}]{MobileNetV2}
Sandler, M.; Howard, A.; Zhu, M.; Zhmoginov, A.; and Chen, L. 2018.
\newblock Mobilenetv2: Inverted Residuals and Linear Bottlenecks.
\newblock In \emph{Proceedings of the IEEE Conference on Computer Vision and
  Pattern Recognition}, 4510--4520.

\bibitem[{Shao and Zhang(2020{\natexlab{a}})}]{bottlenet++}
Shao, J.; and Zhang, J. 2020{\natexlab{a}}.
\newblock Bottlenet++: An End-to-end Approach for Feature Compression in
  Device-edge Co-inference Systems.
\newblock In \emph{IEEE International Conference on Communications Workshops},
  1--6.

\bibitem[{Shao and Zhang(2020{\natexlab{b}})}]{bottlenet2}
Shao, J.; and Zhang, J. 2020{\natexlab{b}}.
\newblock Communication-Computation Trade-Off in Resource-Constrained Edge
  Inference.
\newblock \emph{arXiv preprint arXiv:2006.02166} .

\bibitem[{Simonyan and Zisserman(2015)}]{VGG}
Simonyan, K.; and Zisserman, A. 2015.
\newblock Very deep convolutional networks for large-scale image recognition.
\newblock In \emph{International Conference on Learning Representations}.

\bibitem[{Teerapittayanon, McDanel, and Kung(2016)}]{BranchyNet}
Teerapittayanon, S.; McDanel, B.; and Kung, H. 2016.
\newblock Branchynet: Fast Inference via Early Exiting from Deep Neural
  Networks.
\newblock In \emph{23rd International Conference on Pattern Recognition},
  2464--2469.

\bibitem[{Thulasidasan et~al.(2019)Thulasidasan, Chennupati, Bilmes,
  Bhattacharya, and Michalak}]{MIXUP}
Thulasidasan, S.; Chennupati, G.; Bilmes, J.~A.; Bhattacharya, T.; and
  Michalak, S. 2019.
\newblock On Mixup Training: Improved Calibration and Predictive Uncertainty
  for Deep Neural Networks.
\newblock In \emph{Advances in Neural Information Processing Systems},
  13888--13899.

\bibitem[{Viola and Jones(2001)}]{viola}
Viola, P.; and Jones, M. 2001.
\newblock Rapid Object Detection Using a Boosted Cascade of Simple Features.
\newblock In \emph{Proceedings of the IEEE Conference on Computer Vision and
  Pattern Recognition}, volume~1.

\bibitem[{Wan et~al.(2018)Wan, Wu, Wong, and Lee}]{ConfNet}
Wan, S.; Wu, T.; Wong, W.~H.; and Lee, C. 2018.
\newblock Confnet: Predict with Confidence.
\newblock In \emph{IEEE International Conference on Acoustics, Speech and
  Signal Processing}, 2921--2925.

\bibitem[{Wang et~al.(2018)Wang, Luo, Crankshaw, Tumanov, Yu, and
  Gonzalez}]{IDK}
Wang, X.; Luo, Y.; Crankshaw, D.; Tumanov, A.; Yu, F.; and Gonzalez, J.~E.
  2018.
\newblock IDK Cascades: Fast Deep Learning by Learning not to Overthink.
\newblock In \emph{Conference on Uncertainty in Artificial Intelligence}.

\bibitem[{Zhang, Dalca, and Sabuncu(2019)}]{SD}
Zhang, Z.; Dalca, A.~V.; and Sabuncu, M.~R. 2019.
\newblock Confidence Calibration for Convolutional Neural Networks Using
  Structured Dropout.
\newblock \emph{arXiv preprint arXiv:1906.09551} .

\bibitem[{Zhou, Gao, and Wu(2017)}]{AF}
Zhou, H.; Gao, B.; and Wu, J. 2017.
\newblock Adaptive Feeding: Achieving Fast and Accurate Detections by
  Adaptively Combining Object Detectors.
\newblock In \emph{Proceedings of the IEEE International Conference on Computer
  Vision}, 3505--3513.

\end{thebibliography}

\end{document}